%% file: main.tex
\documentclass[11pt,a4paper]{article}

\usepackage[T1]{fontenc}
\usepackage[utf8]{inputenc}
\usepackage{lmodern}
\usepackage[margin=1in]{geometry}
\usepackage{amsmath,amssymb}
\usepackage{graphicx}
\usepackage{booktabs}
\usepackage{array}          
\usepackage{caption}
\usepackage{subcaption}
\usepackage[numbers,sort&compress]{natbib}
\usepackage{microtype}
\usepackage{xcolor}
\usepackage[hidelinks]{hyperref}
\usepackage{tikz}
\usetikzlibrary{arrows.meta,positioning,fit,backgrounds}

\graphicspath{{figures/}}

\input{tables/values}

\newcommand{\phiage}{\phi_{\mathrm{age}}}
\newcommand{\dphi}{\lvert\Delta\phiage\rvert}
\newcommand{\dpabs}{\lvert\Delta p\rvert}
\newcommand{\discflag}{\texttt{discretize\_\allowbreak continuous}}

\title{\bfseries When Feature Representation Changes Local Explanations:\\
A Controlled Comparison of SHAP and LIME under Numerical Discretization}

\author{%
  Parv Patel \\
  Indian Institute of Technology Palakkad
}
\date{}

\begin{document}
\maketitle

\begin{abstract}
Local feature attributions are computed over a feature representation that an
upstream data-engineering step has chosen, yet they are read as statements
about the underlying quantity. We ask how discretizing one continuous feature
changes local KernelSHAP and LIME explanations when the classifier is held
fixed, and whether the two explainers respond differently. A random forest is
trained on UCI Adult and frozen; the \texttt{age} column is then collapsed onto
equi-width or quantile bin representatives ($k \in \{3,5,10\}$) before the
frozen model receives it, and the same \NInstances{} held-out instances are
explained in every condition across multiple seeds (\NShap{} KernelSHAP and
\NLime{} LIME explanations). We first establish, analytically and by direct
numerical check, that for KernelSHAP this construction is \emph{exactly}
equivalent to explaining the unchanged model at the quantized instance against
a quantized background, so the comparison is between two different effective
inputs rather than a pure explainer effect; for LIME no such equivalence holds.
We then compare each attribution change against a \emph{seed-noise floor}
obtained by re-running the same explainer on the same unchanged condition.
KernelSHAP's age attribution exceeds its own noise floor in all
\ShapNCond{} pre-specified conditions (median ratios \ShapRatioMin{}--%
\ShapRatioMax{}, all significant after Holm correction); on the raw attribution
scale LIME's exceeds its floor in only \LimeNAboveNoise{} of \LimeNCond{}, and
in the rest is no larger than its own run-to-run variation. SHAP's normalized change is
larger than LIME's in every condition (rank-biserial \MagRBMin{}--\MagRBMax{}),
while LIME is \SdRatioMin{}--\SdRatioMax{}$\times$ less consistent across
repeats. Predicted classes change for only \FlipMin{}--\FlipMax{}\% of
instances and held-out ROC-AUC never falls below \AUCWorst{}, yet
\ShapNoFlipMin{}--\ShapNoFlipMax{}\% of instances show a large SHAP attribution
change with no class flip. Feature-importance \emph{rankings}, by contrast, are
barely disturbed (median Spearman $\geq \SpearMin{}$, top-3 overlap
\TopThreeMin{}). The principal limitation is structural: the representation
change and the change in effective input cannot be separated within a
frozen-model design. The contribution is a controlled empirical comparison with
explicit noise and prediction-agreement controls, not a new method or a new
phenomenon.
\end{abstract}

\input{sections/introduction}
\input{sections/related_work}
\input{sections/methodology}
\input{sections/results}
\input{sections/discussion}
\input{sections/limitations}
\input{sections/conclusion}

\section*{Reproducibility}
All code, configuration, saved result tables and figures accompany this
manuscript. Three commands reproduce everything reported here:
\begin{quote}\small\ttfamily
python scripts/run\_all.py -{}-config configs/experiment.yaml -{}-fresh\\
python scripts/verify\_wrapper\_equivalence.py\\
python paper/make\_assets.py
\end{quote}
\noindent
The first re-runs the full pipeline, the second is the equivalence check of
Section~\ref{sec:what-is-explained}, and the third regenerates every table and
inline number in this paper from the saved results. The experiment was
executed with Python 3.13.7,
scikit-learn 1.9.0 \citep{pedregosa2011sklearn}, \texttt{shap} 0.54.1 and
\texttt{lime} 0.2.0.1, and took \Runtime{} minutes on 12 CPU cores.

\bibliographystyle{plainnat}
\bibliography{references}

\end{document}

%% file: tables/values.tex
\newcommand{\AUCDrop}{0.004}

\newcommand{\AUCWorst}{0.9097}
\newcommand{\AgeShiftEWThree}{6.4}
\newcommand{\AgeShiftQTThree}{6.4}
\newcommand{\DpMedMax}{0.0052}
\newcommand{\DpMedMin}{0.0004}
\newcommand{\EWThreeFirstRep}{29.2}
\newcommand{\EquivLimeMax}{1.6e-02}
\newcommand{\EquivLimeRel}{1.5}
\newcommand{\EquivNInst}{12}
\newcommand{\EquivShapMax}{exactly $0$}
\newcommand{\FlipMax}{5.8}
\newcommand{\FlipMin}{0.8}
\newcommand{\HoneVerdict}{supported}
\newcommand{\HthreeVerdict}{supported}
\newcommand{\HtwoVerdict}{supported}
\newcommand{\LimeNAboveNoise}{2}
\newcommand{\LimeNCond}{6}
\newcommand{\LimeRsqBelowHalf}{20.3}
\newcommand{\LimeRsqDisc}{0.620}
\newcommand{\LimeRsqMax}{0.545}
\newcommand{\LimeRsqMed}{0.545}
\newcommand{\LimeRsqMin}{0.543}
\newcommand{\LimeSpearMin}{0.986}
\newcommand{\MagRBMax}{1.00}
\newcommand{\MagRBMin}{0.81}
\newcommand{\MagRatioMax}{14.1}
\newcommand{\MagRatioMin}{3.0}
\newcommand{\NClipped}{3}
\newcommand{\NEncoded}{106}
\newcommand{\NInstances}{120}
\newcommand{\NLime}{5,400}
\newcommand{\NLimeDisc}{3,240}
\newcommand{\NObsChange}{9,240}
\newcommand{\NShap}{3,240}
\newcommand{\NTest}{16,281}
\newcommand{\NTrain}{32,561}
\newcommand{\NoiseLime}{0.00439}
\newcommand{\NoiseLimeHi}{0.00470}
\newcommand{\NoiseLimeLo}{0.00407}
\newcommand{\NoiseLimeRaw}{0.002233}
\newcommand{\NoiseLimeRawHi}{0.002433}
\newcommand{\NoiseLimeRawLo}{0.002014}
\newcommand{\NoiseRatio}{5.1}
\newcommand{\NoiseRatioRaw}{6.8}
\newcommand{\NoiseShap}{0.00087}
\newcommand{\NoiseShapHi}{0.00101}
\newcommand{\NoiseShapLo}{0.00072}
\newcommand{\NoiseShapRaw}{0.000327}
\newcommand{\NoiseShapRawHi}{0.000388}
\newcommand{\NoiseShapRawLo}{0.000255}
\newcommand{\NoiseSpearLime}{0.944}
\newcommand{\NoiseSpearShap}{0.997}
\newcommand{\QTThreeTopHi}{90}
\newcommand{\QTThreeTopLo}{44}
\newcommand{\QTThreeTopRep}{67}
\newcommand{\RankShiftLimeEW}{62}
\newcommand{\RankShiftShapEW}{80}
\newcommand{\RepLimeMed}{0.0005}
\newcommand{\RepLimeMid}{0.0026}
\newcommand{\RepShapMed}{0.0122}
\newcommand{\RepShapMid}{0.0092}
\newcommand{\RhoLimeMed}{0.10}
\newcommand{\RhoShapMax}{0.76}
\newcommand{\RhoShapMin}{0.42}
\newcommand{\Runtime}{62}
\newcommand{\SdRatioMax}{4.6}
\newcommand{\SdRatioMin}{2.5}
\newcommand{\SecAUC}{0.9126}
\newcommand{\SecLime}{0.0072}
\newcommand{\SecN}{60}
\newcommand{\SecShap}{0.0199}
\newcommand{\SeedSweepN}{3}
\newcommand{\SeedSweepShapEW}{0.0517, 0.0516, 0.0516}
\newcommand{\ShapLargeMax}{100}
\newcommand{\ShapLargeMin}{46}
\newcommand{\ShapMedEWThree}{0.0516}
\newcommand{\ShapMedEWThreePct}{5}
\newcommand{\ShapNCond}{6}
\newcommand{\ShapNSig}{6}
\newcommand{\ShapNoFlipMax}{94}
\newcommand{\ShapNoFlipMin}{45}
\newcommand{\ShapRatioMax}{59}
\newcommand{\ShapRatioMin}{2.4}
\newcommand{\ShapReconMax}{2.2e-16}
\newcommand{\SpearMin}{0.979}
\newcommand{\TestAUC}{0.9138}
\newcommand{\TestAcc}{0.8628}
\newcommand{\TestFOne}{0.6713}
\newcommand{\TestPrec}{0.7737}
\newcommand{\TestRec}{0.5928}
\newcommand{\TopThreeMin}{1.00}

%% file: sections/introduction.tex
\section{Introduction}
\label{sec:intro}

A local feature attribution answers the question ``which features drove
\emph{this} prediction?'' by assigning a number to each input feature. The
answer is read as a statement about an underlying quantity---a person's age,
income or occupation---but it is computed over a \emph{representation} of that
quantity that some earlier step in the pipeline chose. Age may be passed as an
integer, bucketed into decades, or standardized; each choice is routine, is
usually made for reasons unrelated to interpretability, and is rarely recorded
alongside the explanation it conditions.

\citet{hwang2025shap} showed that such choices are not innocuous for
SHAP: common data-engineering operations, including binning a continuous
feature, can substantially change SHAP-derived importance even when the trained
classifier is held fixed. If that sensitivity is specific to Shapley-value
estimation, it is an argument for preferring surrogate-based explanations when
the upstream representation is uncertain. If it is shared by LIME, it is a
property of post-hoc local explanation more generally. Addressing this requires
a comparison in which both explainers see identical instances and an identical
prediction target, with each explainer's own settings held fixed across
representations.

Two controls are needed before such a comparison can be interpreted, and both
are easy to omit. First, KernelSHAP and LIME are \emph{stochastic}: KernelSHAP
samples feature coalitions and LIME samples perturbations, so re-running either
on an unchanged input already moves the attributions. An observed
``representation effect'' smaller than that movement is not evidence of
anything. Second, collapsing a feature onto a bin representative changes the
vector the classifier receives. The classifier's parameters may be frozen, but
its predictions need not be, and an attribution that moves because the model
now predicts something different is not the same finding as an attribution that
moves while the prediction stands.

This paper reports a controlled empirical study built around those two
controls. We take the UCI Adult income dataset \citep{becker1996adult}, train a
random forest, freeze it, and discretize the \texttt{age} column with
equi-width and quantile binning at three bin counts. The same \NInstances{}
held-out instances are explained with KernelSHAP and LIME in every condition
and at several random seeds. We measure how far the age attribution moves, how
far the feature ranking moves, how far the prediction moves, and how far each
explainer moves when nothing at all is changed.

\paragraph{Contributions.} We claim the following, and nothing beyond it.

\begin{enumerate}
  \item \textbf{A precise account of what the frozen-model design explains.}
  We show analytically, and confirm numerically to machine precision
  (Section~\ref{sec:what-is-explained}), that for KernelSHAP a frozen model
  behind a quantizing wrapper is \emph{exactly} equivalent to the unchanged
  model explained at the quantized instance against a quantized background.
  The comparison is therefore between two different effective inputs, not a
  pure explainer effect. For LIME the two framings are \emph{not} equivalent,
  because its perturbations are drawn in the raw feature space. This
  distinction is, to our knowledge, not made explicit in the prior work we
  surveyed, and it changes how the resulting numbers may be described.

  \item \textbf{A same-instance SHAP--LIME comparison with an explainer-noise
  floor.} Using the unchanged condition compared against itself across seeds as
  a control, we find that KernelSHAP's age attribution exceeds its own sampling
  noise in all \ShapNCond{} pre-specified conditions, whereas LIME's does so
  in only \LimeNAboveNoise{} on the raw attribution scale. In LIME's remaining
  conditions the change is no larger than its run-to-run variation, which a
  study without this control would not reveal.

  \item \textbf{Evidence that attribution instability and prediction agreement
  are distinct quantities.} Class flips are rare
  (\FlipMin{}--\FlipMax{}\%) and held-out ROC-AUC is essentially unchanged,
  while a large majority of instances nonetheless carry a materially different
  SHAP attribution. We also report a null result: feature-importance
  \emph{rankings} barely move, and for LIME re-running the explainer perturbs
  the ranking more than the discretization does.
\end{enumerate}

We do not claim to be the first to study representation sensitivity, the first
to compare SHAP and LIME, or to introduce a new explanation method. The
phenomenon in contribution~2 was established for SHAP by
\citet{hwang2025shap}; our addition is the same-instance comparison, the noise floor,
the prediction-agreement coupling, and the analysis in contribution~1.

%% file: sections/related_work.tex
\section{Related Work}
\label{sec:related}

\paragraph{The two explainers.} LIME \citep{ribeiro2016lime} explains a
prediction by sampling perturbations around the instance, weighting them by
proximity, and fitting a sparse linear surrogate; the surrogate's coefficients
are the explanation. KernelSHAP \citep{lundberg2017shap} estimates Shapley
values by solving a weighted least-squares problem over feature coalitions,
where an ``absent'' feature is replaced by values drawn from a background
dataset. Both are model-agnostic, both are approximations, and both are
stochastic---facts that matter for any stability claim.
\citet{covert2021kernelshap} analyse KernelSHAP's estimator and show that its
variance at practical sample sizes is large enough to require explicit
treatment, which is one motivation for the noise floor we adopt.

\paragraph{Sensitivity to representation.} The work closest to ours is
\citet{hwang2025shap}, who show that ordinary data-engineering choices---age
binning among them---shift SHAP-derived feature importance, both when the
classifier is retrained on the new representation and, in their attack
experiments, when a trained classifier is held fixed and only the input
representation changes. They note that this could be exploited to make a
discriminatory model appear neutral. They establish the phenomenon for SHAP.
They do not, as far as we can determine, compare it with LIME on identical
instances, nor quantify the effect against the explainer's own sampling noise.
Those are the gaps this paper addresses.

\paragraph{Stability of explanations.} A separate line of work asks whether
explanations are stable under perturbations that \emph{should not} matter.
\citet{alvarezmelis2018robustness} argue that similar inputs ought to produce
similar explanations, propose robustness metrics, and find that common methods
score poorly. \citet{zhang2019lime} characterize the uncertainty in LIME
explanations specifically, attributing it to randomness in the sampling
procedure and to variation across instances; our seed-noise floor is an
empirical measurement of exactly this quantity at our chosen budgets.
\citet{krishna2024disagreement} document a related but distinct problem: different
explanation methods frequently disagree about the same prediction, and
practitioners resolve the disagreement with ad hoc heuristics. Our setting is
narrower---one model, one feature, a controlled change---but the concern is
the same: an explanation consumer has no way to tell a meaningful difference
from an artefact.

\paragraph{Fidelity versus stability.} \citet{yeh2019infidelity} formalize
infidelity and sensitivity as two separate desiderata and show that optimizing
one does not deliver the other. This distinction is directly relevant to our
results: we find that LIME's local surrogate $R^2$ is essentially constant
across conditions in which its attributions nonetheless move, so local fidelity
does not detect the change. We therefore report fidelity as a diagnostic and
not as a validity check.

\paragraph{Feature dependence and off-manifold evaluation.} Both methods
evaluate the model at inputs that need not lie on the data manifold.
KernelSHAP's coalition rows splice the explained instance with background rows,
and LIME perturbs features independently. \citet{aas2021dependent} show that
the standard independence assumption yields inaccurate Shapley values when
features are dependent and propose conditional alternatives.
\citet{chen2020true} frame the same choice as ``true to the model or true to the
data'', arguing that the interventional and observational formulations answer
different questions and that neither is universally correct.
\citet{slack2020fooling} show that the off-manifold behaviour is exploitable:
an adversary can hide a biased model behind benign-looking LIME and SHAP
explanations. We use the libraries' default marginal formulations and treat the
resulting limitation as a scope condition on our conclusions rather than
something we resolve.

\paragraph{The dataset.} UCI Adult \citep{becker1996adult} is a 1994 US census
extract that remains a standard tabular benchmark. \citet{ding2021retiring}
document its limitations---an arbitrary \$50{,}000 threshold, dated
categories, and questionable suitability for fairness conclusions---and propose
replacements. We use it as a methods benchmark for comparing explainers, and
draw no substantive conclusion about income or about people.

\paragraph{Positioning.} Prior work has established (i) that SHAP is sensitive
to feature representation, (ii) that LIME is sensitive to its own sampling,
(iii) that explanation methods disagree with one another, and (iv) that
fidelity and stability are distinct. What we add is a single controlled
experiment in which a representation change, each explainer's own noise, and
the model's prediction agreement are measured simultaneously on the same
instances, so that they can be compared against one another rather than
reported separately.

%% file: sections/methodology.tex
\section{Experimental Setup}
\label{sec:setup}

Table~\ref{tab:setup} summarizes the configuration; this section states the
parts that affect interpretation. Every setting was fixed in a configuration
file before the final results were examined, together with the metrics,
thresholds and seeds.

\begin{table}[t]
\centering
\small
\caption{Experimental configuration. All values are read from the saved run
metadata; the one-hot encoding is applied inside the frozen pipeline, so the
explainers operate on the 12-column interpretable space while the forest
receives \NEncoded{} columns.}
\label{tab:setup}
\input{tables/setup}
\end{table}

\subsection{Data and model}

We use the official UCI Adult train/test partition (\NTrain{} and \NTest{}
rows), which is deterministic and requires no split seed. Two columns are
dropped: \texttt{fnlwgt}, a census sampling weight describing how many people a
row represents rather than an attribute of the person, and
\texttt{education-num}, a deterministic integer re-encoding of
\texttt{education} whose retention would split one piece of information across
two columns and distort every ranking metric. The \texttt{?} placeholder
becomes an explicit \texttt{Missing} category so that no row is dropped and row
identity is preserved across conditions.

The classifier is a random forest (200 trees, minimum leaf size 10,
\texttt{max\_features=sqrt}, seed 42) inside a pipeline whose first stage
one-hot encodes the eight categorical columns into \NEncoded{} columns.
Held-out performance is
ROC-AUC \TestAUC{}, accuracy \TestAcc{}, $F_1$ \TestFOne{}, precision
\TestPrec{} and recall \TestRec{}. The model is then \emph{frozen}: a hash over
every tree's split features, thresholds and leaf values is recorded and
re-checked after all explanations have been computed, and is unchanged. Model
quality was not optimized aggressively; the forest's hyperparameters were
chosen for prediction \emph{speed}, since KernelSHAP's cost is dominated by
model calls.

\subsection{The interpretable feature space}

Explainers and classifier share a single 12-column space in which the four
numeric features keep their natural values and each categorical feature is one
column of integer category codes. One-hot encoding happens inside the pipeline
and is never exposed to the explainers. This matters for two reasons. With \NEncoded{}
one-hot columns exposed, a single real feature such as \texttt{occupation}
would receive 15 separate attributions, and a rank correlation or top-$k$
overlap computed over that space would not correspond to any quantity a reader
interprets. Second, LIME Tabular perturbs one-hot columns independently,
producing rows with two occupations or none; supplying
\texttt{categorical\_features} instead makes LIME resample a whole category
level from its training marginal. That is a mitigation and not a solution---the
marginal is still sampled independently of the other features---and we return
to it in Section~\ref{sec:limitations}.

\subsection{Representation conditions}

Let $x \in \mathbb{R}^{12}$ be an instance and let $q_{s,k}$ map the age
coordinate onto the representative of its bin under scheme
$s \in \{\text{equi-width}, \text{quantile}\}$ with $k$ bins, leaving all other
coordinates unchanged. Bin edges are fitted on the training age column only and
the representative is the bin midpoint. With $f$ the frozen classifier's
positive-class probability, the conditions are
\begin{equation}
  \text{A: } f(x), \qquad
  \text{B/C: } (f \circ q_{s,k})(x), \quad k \in \{3,5,10\}.
  \label{eq:conditions}
\end{equation}
Quantile edges concentrate where the data are dense, so at $k=3$ the top
quantile bin spans \QTThreeTopLo{}--\QTThreeTopHi{} and collapses to
\QTThreeTopRep{}, while equi-width binning sends every instance below age 41.3
to \EWThreeFirstRep{}. The mean absolute age shift at $k=3$ is
\AgeShiftEWThree{} years for equi-width and \AgeShiftQTThree{} for quantile.

\subsection{What the frozen-model design actually explains}
\label{sec:what-is-explained}

Equation~\eqref{eq:conditions} describes a changed \emph{function} evaluated at
an unchanged \emph{instance}. An equally natural reading is an unchanged
function evaluated at a changed instance, $f(q_{s,k}(x))$. Whether these
coincide is not a matter of presentation, and the answer differs between the
two explainers.

\paragraph{KernelSHAP.} Its synthetic coalition rows are coordinate-wise
splices of the explained instance $x$ and a background row $b$: for a coalition
$S$, $\mathrm{splice}(x,b,S)_j = x_j$ if $j \in S$ and $b_j$ otherwise. Since
$q$ acts on a single coordinate, it commutes with the splice,
\begin{equation}
  q\bigl(\mathrm{splice}(x,b,S)\bigr) = \mathrm{splice}\bigl(q(x),q(b),S\bigr),
  \label{eq:commute}
\end{equation}
and the base value $\mathbb{E}_b[f(q(b))]$ is likewise the expectation of $f$
over the quantized background. Hence
\begin{equation}
  \mathrm{KernelSHAP}(f \circ q,\; x,\; B)
  \;=\;
  \mathrm{KernelSHAP}(f,\; q(x),\; q(B))
  \label{eq:equiv}
\end{equation}
exactly, for any background set $B$ and any fixed coalition sample. We verified
\eqref{eq:equiv} numerically on \EquivNInst{} instances in two conditions: the
maximum absolute difference between the two framings is \EquivShapMax{}.

The consequence is important for how the results may be described. For
KernelSHAP, condition B or C \emph{is} the unchanged model explained at a
quantized instance against a quantized background. It is therefore
\textbf{not} a pure explainer effect, and we do not describe it as one. The
frozen classifier guarantees that no parameter changed; it guarantees nothing
about the input the classifier receives.

\paragraph{LIME.} No analogous identity holds. LIME draws perturbations in the
raw feature space using statistics fitted on the raw training matrix, and
regresses the surrogate on those raw perturbations while the labels come from
$f \circ q$. The surrogate is thus a linear fit, in the \emph{un}-quantized
coordinates, to a function that is piecewise constant in age. Explaining
$f$ at $q(x)$ with a training matrix $q(X)$ is a different computation. Running
both on the same instances gives a maximum absolute coefficient difference of
\EquivLimeMax{} (median relative difference up to \EquivLimeRel{}\% of total
attribution mass), confirming non-equivalence.

This asymmetry means the two explainers are not solving the same transformed
problem, and it is one candidate explanation for the divergence reported in
Section~\ref{sec:results}.

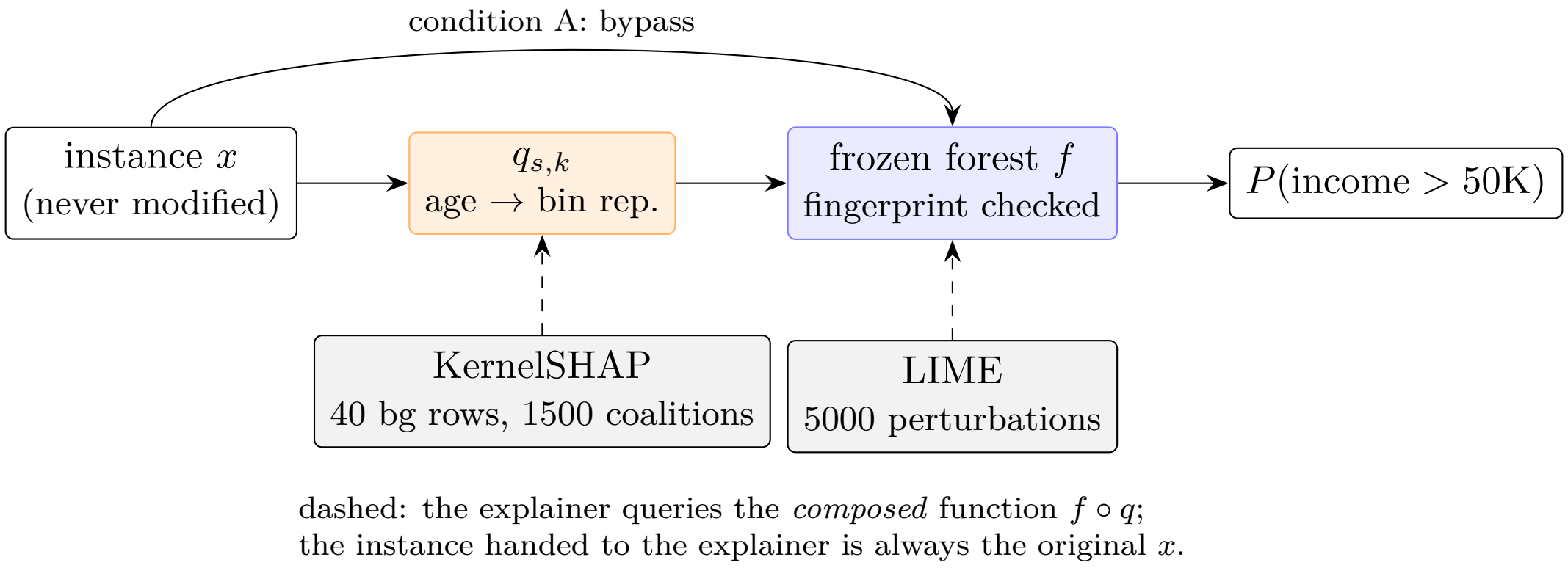
\begin{figure}[t]
\centering
\begin{tikzpicture}[
  node distance=5mm,
  box/.style={draw,rounded corners=2pt,align=center,font=\small,inner sep=4pt},
  frozen/.style={box,fill=blue!8,draw=blue!50},
  quant/.style={box,fill=orange!12,draw=orange!60},
  expl/.style={box,fill=gray!10},
  >={Stealth[length=2mm]}
]
\node[box] (x) {instance $x$\\\footnotesize (never modified)};
\node[quant,right=10mm of x] (q) {$q_{s,k}$\\\footnotesize age $\to$ bin rep.};
\node[frozen,right=10mm of q] (f) {frozen forest $f$\\\footnotesize fingerprint checked};
\node[box,right=10mm of f] (p) {$P(\text{income}>50\mathrm{K})$};
\draw[->] (x) -- (q);
\draw[->] (q) -- (f);
\draw[->] (f) -- (p);
\draw[->] (x.north) .. controls +(0,9mm) and +(0,9mm) .. node[above,font=\scriptsize]
  {condition A: bypass} (f.north);

\node[expl,below=9mm of q] (shap) {KernelSHAP\\\footnotesize 40 bg rows, 1500 coalitions};
\node[expl,below=9mm of f] (lime) {LIME\\\footnotesize 5000 perturbations};
\draw[->,dashed] (shap) -- (q);
\draw[->,dashed] (lime) -- (f);
\node[align=left,font=\scriptsize,below=3mm of shap,xshift=18mm]
  {dashed: the explainer queries the \emph{composed} function $f\circ q$;\\
   the instance handed to the explainer is always the original $x$.};
\end{tikzpicture}
\caption{Experimental design. The classifier's parameters are frozen and the
explained instance is never modified; what changes between conditions is the
effective input the classifier receives. By
Equation~\eqref{eq:equiv} this is \emph{exactly} equivalent, for KernelSHAP, to
explaining the unchanged model at $q(x)$ against a quantized background---so
the comparison is between two effective inputs, not a pure explainer effect.
For LIME the two framings differ, because its perturbations are drawn in the
raw coordinates.}
\label{fig:design}
\end{figure}

\subsection{Explainer configurations}

Both explainers target $P(\text{income}>50\mathrm{K})$ on the same
\NInstances{} test instances, sampled once with a fixed seed and stratified by
the true label (used only for sampling). Explaining the full test set was
infeasible at roughly 2.5\,s per KernelSHAP explanation across nine conditions
and three repeats.

Each explainer's settings are held fixed across every representation
condition, but the two explainers are \emph{not} given equivalent budgets or
reference data. KernelSHAP evaluates the model on 1500 coalitions, each averaged
over 40 background rows; LIME evaluates it on 5000 perturbations drawn from
statistics of the full training partition. The comparison between explainers
is therefore a comparison of two methods at their stated settings, not at a
matched cost.

\paragraph{KernelSHAP.} \texttt{link="identity"}, so attributions are in
probability units and satisfy $\sum_j \phi_j = f(x) - \mathbb{E}[f]$. The
background is 40 real training rows rather than a $k$-means summary: our
categorical columns hold integer codes, and a centroid would be a fractional
code that the one-hot stage maps to an all-zero block, i.e.\ a row with no
occupation. We use 1500 coalition samples and set the $\ell_1$ regularization
to zero. The library default performs AIC-based feature selection and can
return exact zeros, which would create ties in every rank metric and break
additivity; with 12 features and 1500 samples the regression is well
determined without it. Three repeats are run.

\paragraph{LIME.} 5000 perturbations, all 12 features returned, five repeats.
Critically, we set \discflag{} to \texttt{False}. The library
\emph{default} is \texttt{True}, which bins every continuous feature into
quartiles and fits the surrogate on bin-membership indicators. Leaving it
enabled would mean age is already discretized by LIME in both the raw and the
discretized condition---the manipulation under study applied to our own
control. We run the default separately as a sensitivity condition
(\NLimeDisc{} further explanations). A consequence of disabling it is that a
LIME coefficient here is a local slope per training standard deviation, not an
attribution in probability units.

\paragraph{Seeding.} Both explainers are reseeded per (repeat, instance) from a
deterministic derivation, so each explanation is reproducible independently of
how instances were batched across parallel workers. The seed depends on the
repeat and the instance but not on the condition. Within a repeat, condition A
and each discretized condition therefore use the same KernelSHAP coalitions and
the same LIME perturbations, and differ only in the function being explained.
The seed-noise floor, by contrast, compares two different seeds. This
asymmetry matters when a representation effect is compared with the floor, and
we note it wherever such a comparison is interpreted.

\subsection{Metrics}

For each instance, each explainer and each repeat we compare a discretized
condition against condition A at the \emph{same} seed. We report the absolute
change in the age attribution, both raw and normalized as
$\phi_j / \sum_i |\phi_i|$; the change in age's rank by $|\phi|$; the Spearman
correlation between the two $|\phi|$ orderings; and top-$k$ overlap for
$k \in \{1,3,5\}$. Normalization is the only form in which SHAP and LIME
magnitudes are compared, since SHAP values are in probability units while LIME
coefficients are per training standard deviation; rank-based metrics are
already scale-free. Raw values are retained in the released tables.

\paragraph{Seed-noise floor.} Condition A is compared against \emph{itself}
across every distinct pair of seeds, giving the attribution movement that
repetition alone produces at these budgets. This is the reference against which
every attribution change is read, and the 95th percentile of this distribution
defines ``large'' for the purposes of Section~\ref{sec:results}.

\paragraph{Prediction agreement.} For each condition we record the class-flip
rate, the distribution of $\dpabs$, and held-out ROC-AUC under the quantized
input. Per-instance $\dpabs$ and the flip indicator are joined onto every
attribution comparison, so the two can be examined together.

\paragraph{Explanation quality.} For KernelSHAP we report the reconstruction
error $|\mathbb{E}[f] + \sum_j \phi_j - f(x)|$; for LIME, the weighted local
$R^2$ of its surrogate. These are not interchangeable and neither is a validity
check, a point we return to in Sections~\ref{sec:results}
and~\ref{sec:limitations}.

\subsection{Statistical analysis}

Instances are the independent unit; the several rows per instance (one per
repeat) are not. Confidence intervals come from a cluster bootstrap that
resamples whole instances (2000 resamples). Paired comparisons use the Wilcoxon
signed-rank test on \emph{one value per instance} (the median across repeats),
reported with the rank-biserial correlation as an effect size. Each hypothesis
is tested once per (condition, explainer), and Holm's step-down procedure
\citep{holm1979simple} controls the family-wise error rate within each family;
Holm is used because it does not assume independence, which these tests do not
have---they share instances and a common reference condition. A test that
cannot run leaves a verdict \emph{inconclusive} rather than \emph{not
supported}.

%% file: tables/setup.tex
\begin{tabular}{@{}l >{\raggedright\arraybackslash}p{0.60\linewidth}@{}}
\toprule
Component & Setting \\
\midrule
Dataset & UCI Adult (Becker \& Kohavi, 1996) \\
Split & official UCI train/test partition \\
Training / test rows & 32,561 / 16,281 \\
Positive rate (train / test) & 0.241 / 0.236 \\
Features (interpretable / encoded) & 12 / 106 \\
Dropped columns & \texttt{fnlwgt}, \texttt{education-num} \\
Classifier & random forest, 200 trees, min.\ leaf 10, \texttt{max\_features=sqrt} \\
Model seed & 42 \\
Evaluation instances & 120 (stratified) \\
KernelSHAP & 40 background rows, nsamples $=$ 1500, $\ell_1$ reg.\ $=0$, 3 repeats \\
LIME & 5000 perturbations, all 12 features, \texttt{discretize\_continuous=False}, 5 repeats \\
Explanations computed & 3,240 SHAP $+$ 5,400 LIME ($+$ 3,240 LIME sensitivity) \\
\bottomrule
\end{tabular}

%% file: sections/results.tex
\section{Results}
\label{sec:results}

\subsection{Explainer noise is not negligible, and differs between methods}
\label{sec:noise}

Re-running the \emph{unchanged} condition at a different seed already moves the
age attribution. In raw attribution units the median movement is
\NoiseShapRaw{} (95\% CI [\NoiseShapRawLo{}, \NoiseShapRawHi{}]) for
KernelSHAP and \NoiseLimeRaw{} [\NoiseLimeRawLo{}, \NoiseLimeRawHi{}] for
LIME---a factor of \NoiseRatioRaw{} on the raw scale. After normalizing to
shares of total attribution mass the figures are
\NoiseShap{} [\NoiseShapLo{}, \NoiseShapHi{}] and \NoiseLime{}
[\NoiseLimeLo{}, \NoiseLimeHi{}], so LIME is \NoiseRatio{} times noisier on
the normalized scale at these budgets. The same holds for rankings---median
Spearman correlation between two repeats is \NoiseSpearShap{} for SHAP and
\NoiseSpearLime{} for LIME.

\paragraph{Units.} Two scales appear below and they are not interchangeable.
Comparisons of an explainer \emph{against itself} (H1, and the ``large
change'' threshold used for H3) are made in \emph{raw} attribution units,
where no normalization is needed or desirable. Comparisons \emph{between} SHAP
and LIME (H2) use the \emph{normalized} attribution, because SHAP values are in
probability units and LIME coefficients are per training standard deviation.
Each table states which scale it uses.

\subsection{H1: the age attribution moves, reliably for SHAP only}

\begin{table}[t]
\centering
\small
\caption{\textbf{H1.} Median per-instance change
$\widetilde{\Delta} = \mathrm{med}\,\dphi$ in the age attribution under each
discretization, in \emph{raw} attribution units. H1 compares each explainer
against \emph{itself}, so no cross-method normalization is applied here; the
normalized scale is used for the SHAP--LIME comparison in Table~\ref{tab:h2}.
Columns give the ratio to the same explainer's seed-noise floor, the
rank-biserial effect size $r$ of a paired Wilcoxon test against that floor, and
the Holm-corrected $p$. Ratios above 1 (bold) indicate the representation
effect exceeds the explainer's own sampling noise; a negative $r$ with a small
$p$ means the effect is smaller than that noise, and an $r$ near zero with a
large $p$ (LIME, QT $k{=}5$) means the two cannot be distinguished. Each
discretized condition is compared with condition A at the same seed, so the two
share the explainer's random draws; the floor compares different seeds. A ratio
below 1 therefore means the representation effect is smaller than run-to-run
variation, not that it is absent. The final row is the median
across instances of each instance's median change across seed pairs, which is
the quantity the paired test compares against and differs slightly from the
pooled median quoted in Section~\ref{sec:noise}.
EW\,$=$\,equi-width, QT\,$=$\,quantile.}
\label{tab:h1}
\input{tables/h1}
\end{table}

\begin{figure}[t]
\centering
\includegraphics[width=\linewidth]{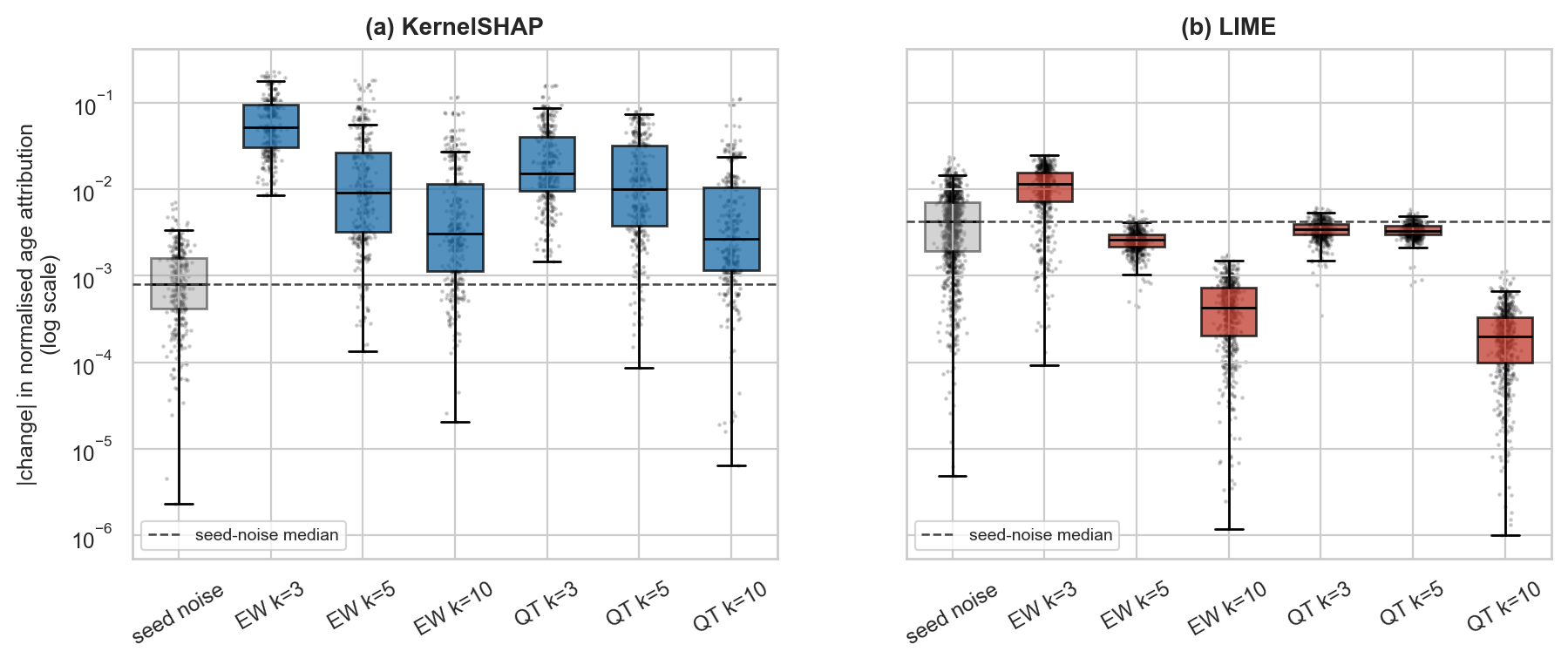}
\caption{Per-instance absolute change in the normalized age attribution between
the original representation and each discretization, by explainer, on a
\emph{shared logarithmic} axis. Grey boxes and dashed lines mark the
seed-noise floor (the same comparison between two repeats of the unchanged
condition). Boxes give median and interquartile range; dots are individual
instance--repeat observations. Every KernelSHAP condition sits above its floor;
on this normalized scale every LIME condition except EW $k{=}3$ sits at or
below its own. On the raw scale used for the H1 tests (Table~\ref{tab:h1}),
LIME's QT $k{=}3$ median lies slightly above its floor instead. Values below
$10^{-6}$ (\NClipped{} of \NObsChange{} observations) are clipped to the axis
floor.}
\label{fig:attrchange}
\end{figure}

Table~\ref{tab:h1} compares each condition against the matching noise floor.
KernelSHAP exceeds its floor in all \ShapNCond{} pre-specified conditions, by
median ratios of \ShapRatioMin{} to \ShapRatioMax{}, with all \ShapNSig{} of
\ShapNCond{} tests significant after Holm correction and large positive effect
sizes. On this raw scale LIME
exceeds its floor in only \LimeNAboveNoise{} of \LimeNCond{} conditions, both
at the coarsest binning ($k=3$). At QT $k{=}5$ its response is
indistinguishable from the floor (ratio close to 1, $r$ close to zero, and the
only non-significant H1 test after Holm correction). In the remaining three
conditions (EW $k{=}5$, EW $k{=}10$ and QT $k{=}10$) its response is
significantly \emph{smaller} than the variation between two of its own runs.
Because the representation comparison shares the explainer's random draws
(Section~\ref{sec:setup}), this says the effect is small relative to
run-to-run variation; it does not say the effect is absent.
Figure~\ref{fig:attrchange} shows the full per-instance distributions on a
shared logarithmic axis, on the \emph{normalized} scale; there LIME's QT
$k{=}3$ median falls just below its floor, so the count of LIME conditions
above the floor depends on the scale and should be read with that caveat.

The verdict for H1 is \textbf{\HoneVerdict{}} as the hypothesis is stated---it
asserts a change ``for at least some instances''---but the qualifier carries
real weight for LIME. A study reporting LIME's raw attribution change without a
noise floor would have described the $k \geq 5$ conditions as positive results
without noting that, in those conditions, the change is no larger than the
variation produced by re-running LIME on unchanged input.

\subsection{Rankings barely move: a null result}

\begin{table}[t]
\centering
\small
\caption{Ranking stability. Median Spearman correlation between the original
and discretized $|\phi|$ orderings, median top-3 overlap, and median absolute
shift in the age feature's own rank. The final row is the seed-noise floor.
Note that for LIME the ranking agreement under representation change is
\emph{higher} than between two repeats of the unchanged condition; the
representation comparison is made at a shared seed, the floor across seeds.}
\label{tab:ranking}
\input{tables/ranking}
\end{table}

\begin{figure}[t]
\centering
\includegraphics[width=\linewidth]{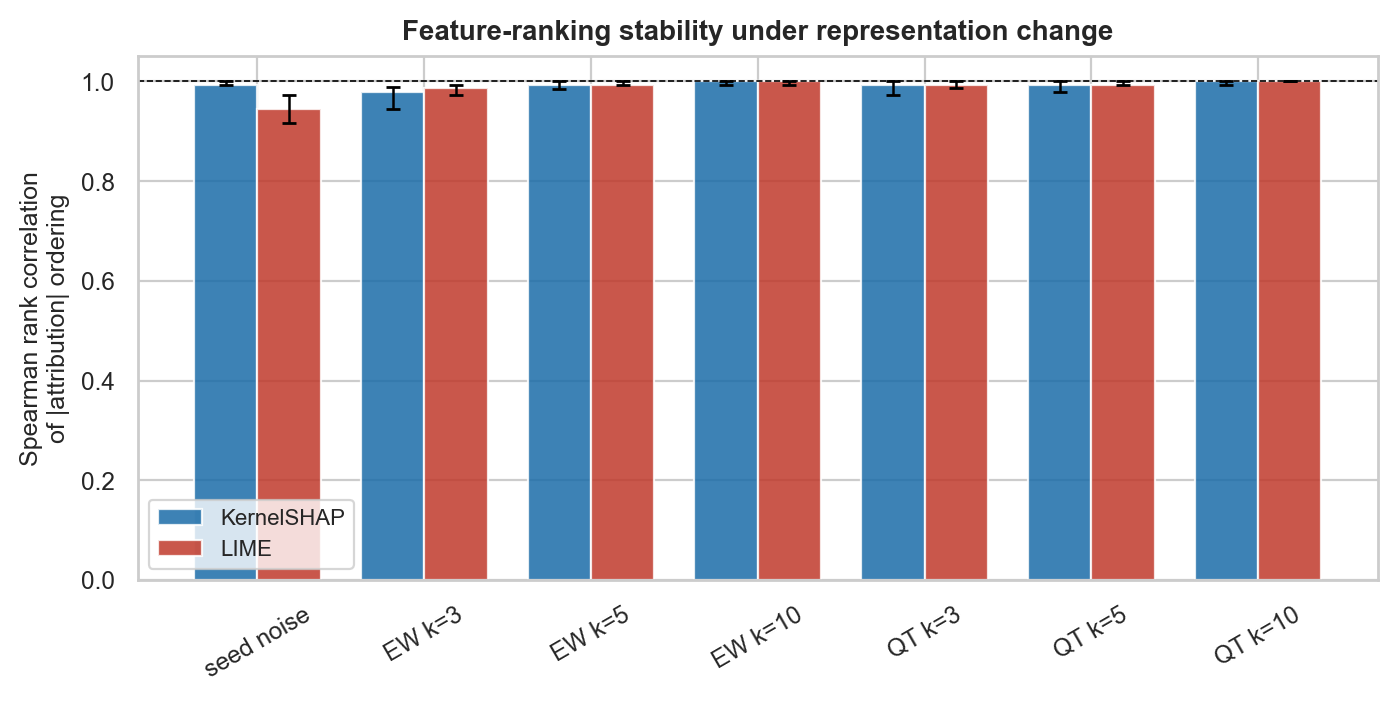}
\caption{Median Spearman rank correlation between the feature-importance
ordering under the original representation and under each discretization, with
interquartile range; 1 means the ordering is unchanged. The leftmost pair is
the seed-noise floor, the agreement between two repeats of the unchanged
condition. The discretized conditions are compared with condition A at a shared
seed, so they can show higher agreement than the floor, as LIME does here.}
\label{fig:ranking}
\end{figure}

Median Spearman correlation between the original and discretized importance
orderings never falls below \SpearMin{}, and median top-1, top-3 and top-5
overlap is \TopThreeMin{} in every condition (Table~\ref{tab:ranking};
Figure~\ref{fig:ranking} plots the Spearman values with their spread). No
condition disturbs the top of the ranking. The age feature's own rank does
move---by at least one position for \RankShiftShapEW{}\% of instances under
SHAP and \RankShiftLimeEW{}\% under LIME at EW $k{=}3$---but the ordering as a
whole is robust.

The comparison with the noise floor is sharper still: for LIME, ranking
agreement \emph{under representation change} ($\geq \LimeSpearMin{}$) exceeds its
agreement between two repeats of the \emph{unchanged} condition
(\NoiseSpearLime{}). Re-running LIME perturbs the importance ordering more than
discretizing the feature does. This comparison should not be read as
discretization making LIME more stable: each discretized condition is compared
with condition A at the same seed, so the two explanations share LIME's
perturbations, whereas the floor compares two different seeds. What it shows is
that the effect of discretization on the ranking is smaller than LIME's
run-to-run variation. This is a null result with respect to ranking, and we
report it as one.

\subsection{H2: SHAP responds more, LIME varies more}

\begin{table}[t]
\centering
\small
\caption{\textbf{H2.} SHAP versus LIME on the same instances in each
condition. Left: median
per-instance age-attribution change on the \emph{normalized} scale
($\phi_j/\sum_i|\phi_i|$), with the rank-biserial effect size and
Holm-corrected $p$ of a paired Wilcoxon test between the two explainers on
per-instance medians. Right: each explainer's median within-instance standard
deviation across repeats, a measure of its own consistency.}
\label{tab:h2}
\input{tables/h2}
\end{table}

Table~\ref{tab:h2} shows that SHAP's normalized response exceeds LIME's in
every condition, by factors of \MagRatioMin{} to \MagRatioMax{}, with
rank-biserial correlations between \MagRBMin{} and \MagRBMax{} and all
Holm-corrected $p$ values below 0.001. In the opposite direction, LIME is the
less consistent method: its median within-instance standard deviation across
repeats is \SdRatioMin{}--\SdRatioMax{}$\times$ SHAP's. H2 is therefore
\textbf{\HtwoVerdict{}}. The magnitude difference is supported by the paired
tests above; the consistency difference is descriptive, since no significance
test was applied to it.

An alternative account of the magnitude difference deserves stating, because
our design cannot rule it out. Shapley values must sum to
$f(x) - \mathbb{E}[f]$, so any change in the model output is necessarily
redistributed across features; a LIME coefficient is a local slope that can
remain unchanged while the surrogate's intercept absorbs the shift. Part of
SHAP's larger response may therefore be definitional rather than evidence of
greater sensitivity. The correlations in Section~\ref{sec:h3} are consistent
with this reading. A second candidate is the non-equivalence established in
Section~\ref{sec:what-is-explained}: LIME is fitting a linear surrogate in
\emph{un}-quantized coordinates to a function that is piecewise constant in
age, which may smooth the step away. We cannot adjudicate between these
accounts here.

\subsection{H3: explanations move while predictions do not}
\label{sec:h3}

\begin{table}[t]
\centering
\small
\caption{\textbf{H3} and prediction agreement. Left: fraction of evaluation
instances whose predicted class flips, median $\dpabs$, and held-out ROC-AUC under
the quantized input. Centre: fraction of instances whose age attribution moves
beyond the 95th percentile of the explainer's own seed-noise distribution
(raw units) \emph{while the predicted class is unchanged}. Right: Spearman correlation between $\dphi$ and $\dpabs$
across instances.}
\label{tab:predh3}
\input{tables/prediction_h3}
\end{table}

\begin{figure}[t]
\centering
\includegraphics[width=\linewidth]{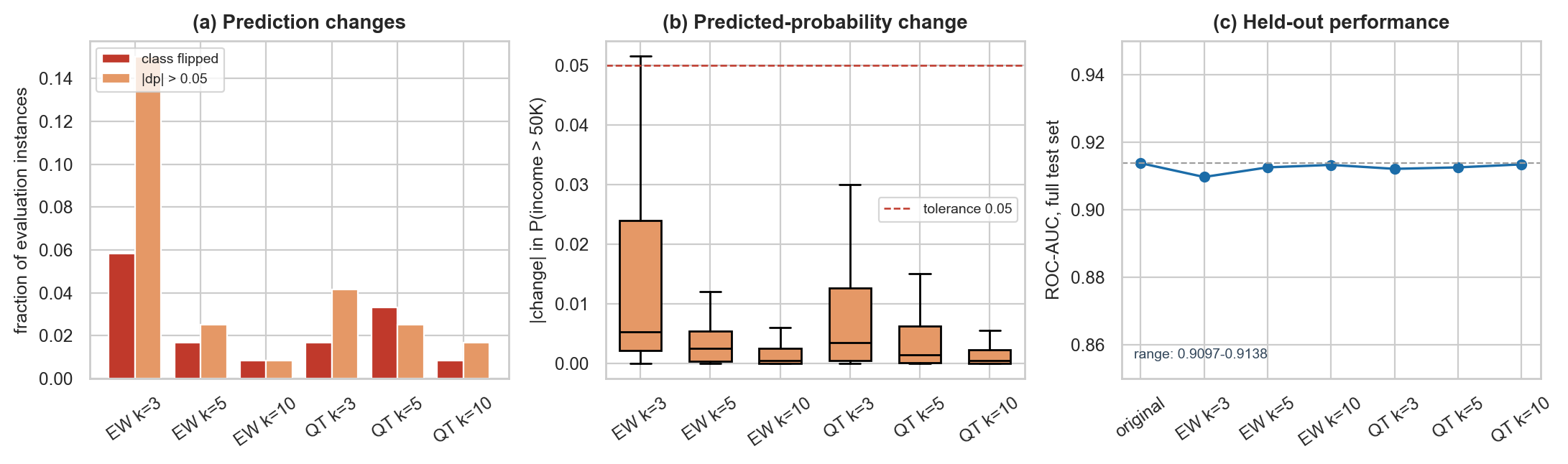}
\caption{Prediction-behaviour diagnostics for the frozen classifier under each
discretization. (a) Fraction of evaluation instances whose predicted class
changes, and whose predicted probability moves by more than the pre-declared
tolerance of 0.05. (b) Distribution of $\dpabs$. (c) ROC-AUC on the full held-out
test set under the quantized input; the axis is fixed to 0.85--0.95 so that the
near-constancy is not visually exaggerated.}
\label{fig:predagree}
\end{figure}

Quantizing age barely disturbs the classifier. Predicted classes change for
\FlipMin{}--\FlipMax{}\% of evaluation instances, median $\dpabs$ lies between
\DpMedMin{} and \DpMedMax{}, and held-out ROC-AUC never falls below
\AUCWorst{} against \TestAUC{} for the original representation
(Table~\ref{tab:predh3}, Figure~\ref{fig:predagree}). By any conventional
validation check, the pipeline is unaffected.

\begin{figure}[t]
\centering
\includegraphics[width=\linewidth]{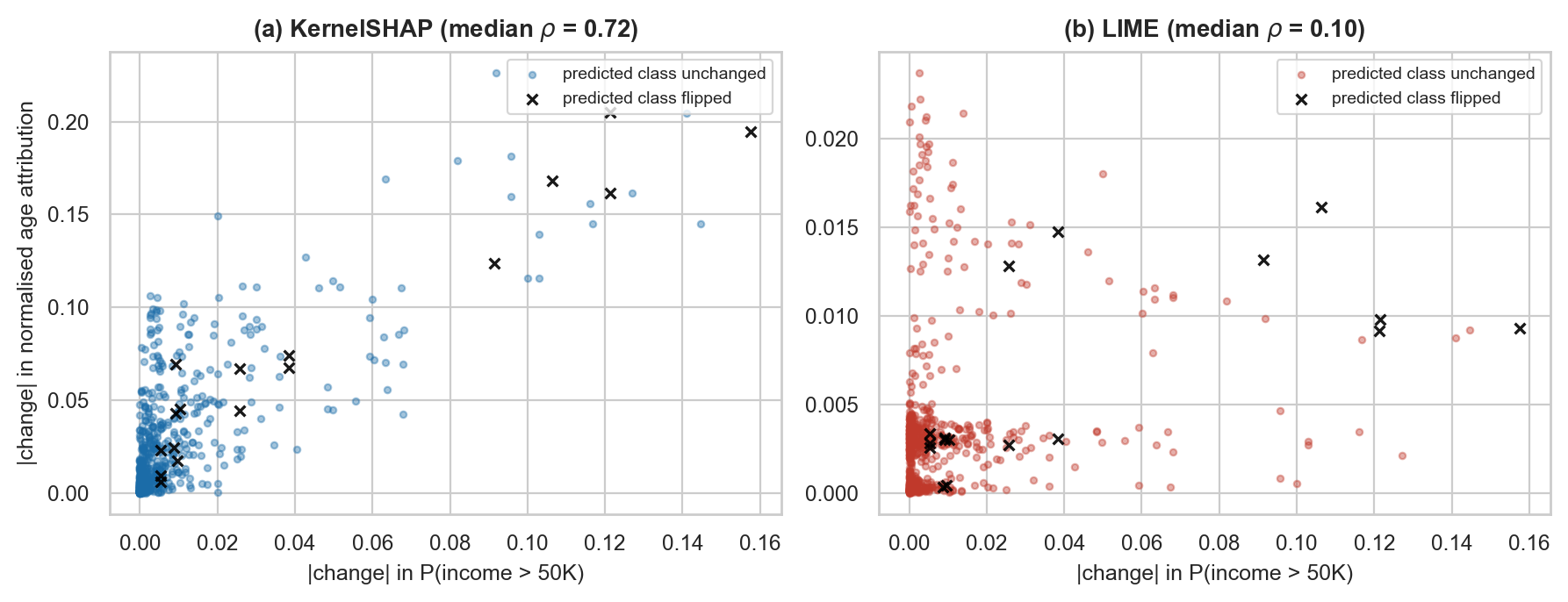}
\caption{Per-instance absolute change in the normalized age attribution against
the absolute change in predicted probability, pooled over the six
pre-specified conditions; crosses mark instances whose predicted class
flipped. The dense vertical band at $\dpabs \approx 0$ is the H3 result:
explanations that moved substantially while the prediction did not. The quoted
$\rho$ is the median across conditions of the within-condition Spearman
correlation.}
\label{fig:attrvspred}
\end{figure}

Yet \ShapLargeMin{}--\ShapLargeMax{}\% of instances show a SHAP age-attribution
change exceeding the 95th percentile of SHAP's own seed-noise distribution, and
\ShapNoFlipMin{}--\ShapNoFlipMax{}\% do so \emph{with no class flip at all}.
The two quantities are related but far from identical: the Spearman correlation
between $\dphi$ and $\dpabs$ is \RhoShapMin{}--\RhoShapMax{} for SHAP and about
\RhoLimeMed{} for LIME (Figure~\ref{fig:attrvspred}). H3 is
\textbf{\HthreeVerdict{}}.

We emphasize what this does and does not show. It does not show that the
explanation moved \emph{without cause}: by
Section~\ref{sec:what-is-explained}, the effective input genuinely changed, and
the moderate positive correlation for SHAP indicates that part of the movement
tracks the prediction. What it shows is that the two cannot substitute for one
another---prediction agreement near 100\% coexists with widespread attribution
change, so verifying the former tells a practitioner little about the latter.

\subsection{Explanation quality diagnostics}

\begin{table}[t]
\centering
\small
\caption{Explanation-quality diagnostics. KernelSHAP's reconstruction error is
at machine precision by construction with $\ell_1$ regularization disabled.
LIME's weighted local $R^2$ is essentially constant across conditions whose
attributions differ, and rises when LIME's own internal discretization is
re-enabled (final column).}
\label{tab:quality}
\input{tables/quality}
\end{table}

KernelSHAP's reconstruction error is at most \ShapReconMax{} across all
\NShap{} explanations, confirming that the efficiency axiom holds---but it does
so by construction and carries no information about stability.

\begin{figure}[t]
\centering
\includegraphics[width=\linewidth]{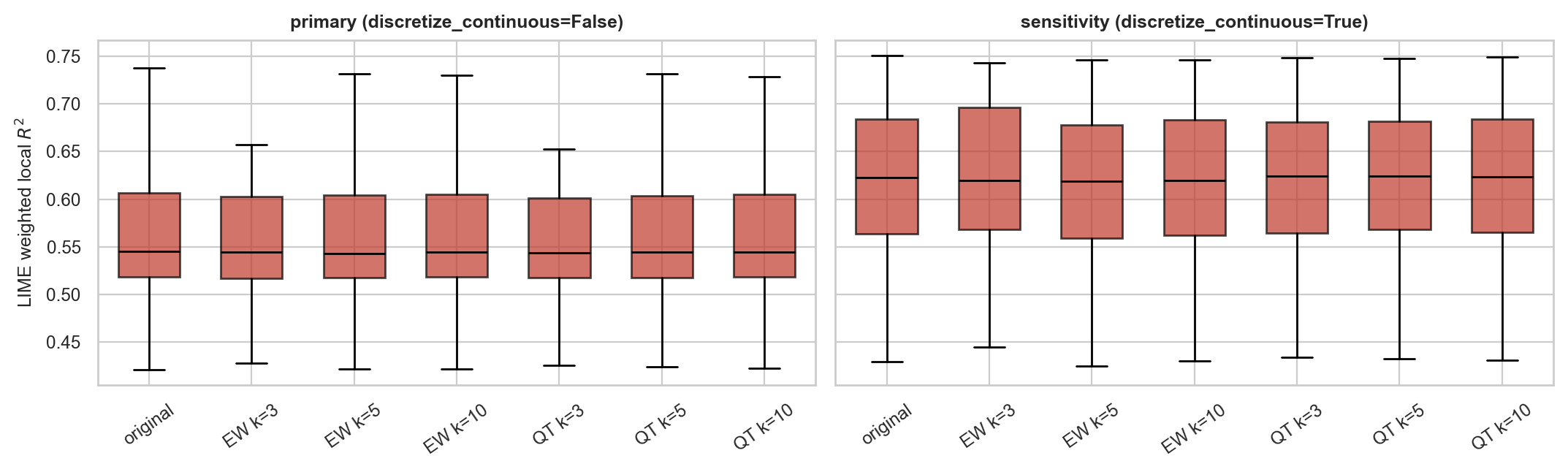}
\caption{Distribution of LIME's weighted local surrogate $R^2$ across
evaluation instances and seeds. Left: the primary configuration, with LIME's
internal discretization disabled so the only discretization present is the one
under study. Right: the sensitivity run with the library default re-enabled. A
high local $R^2$ means the surrogate fits the model well in the sampled
neighbourhood; it does not imply the explanation is stable across seeds or
across representations.}
\label{fig:fidelity}
\end{figure}

LIME's median weighted local $R^2$ is \LimeRsqMed{} for the original
representation and varies only between \LimeRsqMin{} and \LimeRsqMax{} across
all conditions, with \LimeRsqBelowHalf{}\% of explanations below 0.5
(Table~\ref{tab:quality}, Figure~\ref{fig:fidelity}). The median is therefore
close to constant across conditions, although individual explanations' $R^2$
values do differ between conditions. Median local fidelity thus barely responds
to the manipulation, including at EW $k{=}3$, where LIME's age attribution
moved beyond its own noise floor. When LIME's own internal
discretization is switched back on, median $R^2$ rises to \LimeRsqDisc{}---a
better local fit, obtained by explaining binned features rather than raw ones.
This is a concrete instance of the fidelity/stability distinction of
\citet{yeh2019infidelity}.

\subsection{Sensitivity analyses}

\begin{figure}[t]
\centering
\includegraphics[width=\linewidth]{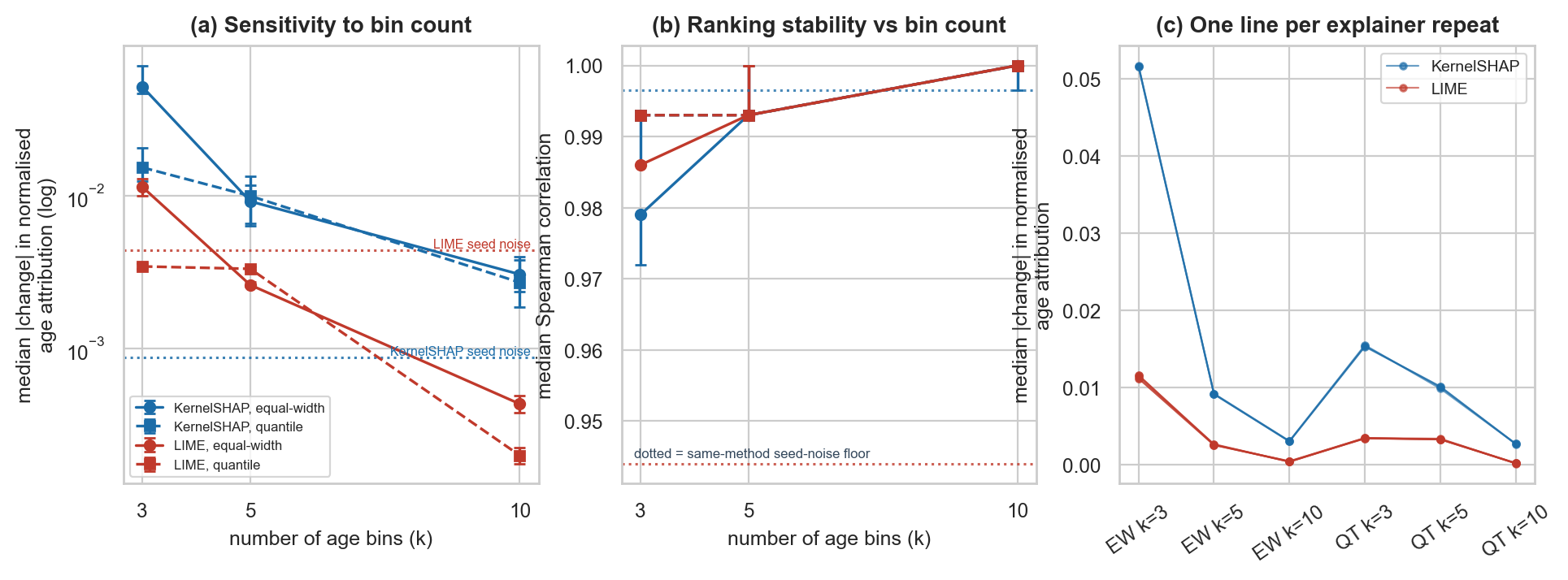}
\caption{Sensitivity analyses. (a) Median normalized attribution change versus
bin count, with 95\% cluster-bootstrap intervals; dotted horizontal lines are
each method's own seed-noise floor. LIME's equi-width response falls below its
floor from $k=5$ onward and its quantile response lies below it at every bin
count, while KernelSHAP's stays above at every bin count. (b) Ranking stability
versus bin count. LIME sits above its noise floor at every bin count, so
discretization perturbs its ordering less than re-running it does; KernelSHAP
falls slightly below its own floor at $k=3$ and $k=5$ and is above it at
$k=10$. (c) The headline metric recomputed within each individual
repeat; the per-seed lines coincide almost exactly.}
\label{fig:sensitivity}
\end{figure}

Figure~\ref{fig:sensitivity} summarizes three sweeps. The effect decays
monotonically with bin count for both schemes and both explainers. On this
normalized scale, LIME's equi-width curve crosses below its own noise floor
between $k=3$ and $k=5$ and its quantile curve lies below the floor at every
$k$, while SHAP's remains above throughout. For ranking stability the picture
is different: LIME stays above its floor at every $k$, whereas SHAP falls
slightly below its own floor at $k=3$ and $k=5$.
Recomputing the headline metric within each individual repeat gives
near-coincident lines (SHAP at EW $k{=}3$: \SeedSweepShapEW{} across \SeedSweepN{}
seeds), so no conclusion depends on a single seed. Replacing the bin midpoint
with the bin's training median changes magnitudes (SHAP at EW $k{=}5$:
$\RepShapMid{} \to \RepShapMed{}$; LIME: $\RepLimeMid{} \to \RepLimeMed{}$)
without changing the ordering of the two explainers.

In the \emph{secondary} experiment a second forest is trained on
quantile-discretized data (held-out ROC-AUC \SecAUC{} against \TestAUC{}) and
explained on discretized inputs over \SecN{} instances. Median normalized
attribution change is \SecShap{} for SHAP and \SecLime{} for LIME---the same
ordering, at roughly twice the frozen-model $k{=}5$ magnitude. This condition
changes the model \emph{and} the representation simultaneously and is reported
as a contrast, not as a decomposition.

\subsection{Statistical versus practical significance}

All H2 magnitude tests and all but one H1 test on the pre-specified conditions
are significant after Holm correction, most with $p$ below $10^{-3}$ and
rank-biserial correlations close to $\pm 1$. The exception is LIME at QT
$k{=}5$, whose response cannot be distinguished from its noise floor
(Table~\ref{tab:h1}). Significance here is cheap: with \NInstances{} paired
instances and
highly consistent within-instance behaviour, even small systematic differences
are detectable. The practically meaningful observations are different in
character, and we separate them deliberately: that SHAP's median response at
EW $k{=}3$ is \ShapMedEWThree{}, i.e.\ about \ShapMedEWThreePct{}\% of total
attribution mass (large enough to reorder
mid-ranking features), that LIME's response at $k \geq 5$ is no larger than
its own re-run variability (so a user comparing two single LIME runs could not
separate it from noise), and that the top of the feature
ranking never moves (so a reader who consults only the top-3 features would see
nothing).

%% file: tables/h1.tex
\begin{tabular}{@{}l rrrr rrrr@{}}
\toprule
 & \multicolumn{4}{c}{KernelSHAP} & \multicolumn{4}{c}{LIME} \\
\cmidrule(lr){2-5}\cmidrule(lr){6-9}
Condition & $\widetilde{\Delta}$ & ratio & $r$ & $p_{\mathrm{Holm}}$ & $\widetilde{\Delta}$ & ratio & $r$ & $p_{\mathrm{Holm}}$ \\
\midrule
EW $k{=}3$ & 0.0215 & \textbf{59.43} & +1.00 & $<$0.001 & 0.0073 & \textbf{3.23} & +0.94 & $<$0.001 \\
EW $k{=}5$ & 0.0033 & \textbf{9.06} & +0.93 & $<$0.001 & 0.0013 & 0.59 & -0.87 & $<$0.001 \\
EW $k{=}10$ & 0.0012 & \textbf{3.43} & +0.77 & $<$0.001 & 0.0001 & 0.06 & -1.00 & $<$0.001 \\
QT $k{=}3$ & 0.0062 & \textbf{17.02} & +1.00 & $<$0.001 & 0.0026 & \textbf{1.15} & +0.35 & 0.002 \\
QT $k{=}5$ & 0.0036 & \textbf{9.89} & +0.99 & $<$0.001 & 0.0022 & 0.99 & -0.04 & 0.681 \\
QT $k{=}10$ & 0.0009 & \textbf{2.38} & +0.70 & $<$0.001 & 0.0002 & 0.10 & -1.00 & $<$0.001 \\
\midrule
\emph{seed-noise floor} & \emph{0.00036} & \multicolumn{3}{c}{} & \emph{0.00226} & \multicolumn{3}{c}{} \\
\bottomrule
\end{tabular}

%% file: tables/ranking.tex
\begin{tabular}{@{}l rr rr rr@{}}
\toprule
 & \multicolumn{2}{c}{Spearman $\rho$} & \multicolumn{2}{c}{top-3 overlap} & \multicolumn{2}{c}{age rank shift} \\
\cmidrule(lr){2-3}\cmidrule(lr){4-5}\cmidrule(lr){6-7}
Condition & SHAP & LIME & SHAP & LIME & SHAP & LIME \\
\midrule
EW $k{=}3$ & 0.979 & 0.986 & 1.00 & 1.00 & 1.50 & 1.00 \\
EW $k{=}5$ & 0.993 & 0.993 & 1.00 & 1.00 & 0.00 & 0.00 \\
EW $k{=}10$ & 1.000 & 1.000 & 1.00 & 1.00 & 0.00 & 0.00 \\
QT $k{=}3$ & 0.993 & 0.993 & 1.00 & 1.00 & 0.00 & 0.00 \\
QT $k{=}5$ & 0.993 & 0.993 & 1.00 & 1.00 & 0.00 & 0.00 \\
QT $k{=}10$ & 1.000 & 1.000 & 1.00 & 1.00 & 0.00 & 0.00 \\
\midrule
\emph{seed noise} & \emph{0.997} & \emph{0.944} & \emph{1.00} & \emph{1.00} & \emph{0.00} & \emph{0.75} \\
\bottomrule
\end{tabular}

%% file: tables/h2.tex
\begin{tabular}{@{}l rr rr rr@{}}
\toprule
 & \multicolumn{4}{c}{Size of change $\widetilde{\Delta}_{\mathrm{norm}}$} & \multicolumn{2}{c}{Within-instance s.d.} \\
\cmidrule(lr){2-5}\cmidrule(lr){6-7}
Condition & SHAP & LIME & $r$ & $p_{\mathrm{Holm}}$ & SHAP & LIME \\
\midrule
EW $k{=}3$ & 0.0516 & 0.0115 & +1.00 & $<$0.001 & 0.00024 & 0.00087 \\
EW $k{=}5$ & 0.0092 & 0.0026 & +0.87 & $<$0.001 & 0.00009 & 0.00038 \\
EW $k{=}10$ & 0.0031 & 0.0005 & +0.94 & $<$0.001 & 0.00005 & 0.00021 \\
QT $k{=}3$ & 0.0154 & 0.0035 & +0.99 & $<$0.001 & 0.00016 & 0.00040 \\
QT $k{=}5$ & 0.0101 & 0.0033 & +0.81 & $<$0.001 & 0.00010 & 0.00029 \\
QT $k{=}10$ & 0.0027 & 0.0002 & +0.99 & $<$0.001 & 0.00004 & 0.00012 \\
\bottomrule
\end{tabular}

%% file: tables/prediction_h3.tex
\begin{tabular}{@{}l rrr rr rr@{}}
\toprule
 & \multicolumn{3}{c}{Prediction behaviour} & \multicolumn{2}{c}{Large $|\Delta\phi_{\mathrm{age}}|$, no flip} & \multicolumn{2}{c}{$\rho(|\Delta\phi|,|\Delta p|)$} \\
\cmidrule(lr){2-4}\cmidrule(lr){5-6}\cmidrule(lr){7-8}
Condition & flip & $\widetilde{|\Delta p|}$ & AUC & SHAP & LIME & SHAP & LIME \\
\midrule
EW $k{=}3$ & 5.8\% & 0.0052 & 0.9097 & 94\% & 55\% & +0.71 & +0.16 \\
EW $k{=}5$ & 1.7\% & 0.0024 & 0.9125 & 73\% & 0\% & +0.74 & +0.04 \\
EW $k{=}10$ & 0.8\% & 0.0005 & 0.9133 & 52\% & 0\% & +0.73 & -0.16 \\
QT $k{=}3$ & 1.7\% & 0.0034 & 0.9121 & 90\% & 0\% & +0.42 & -0.06 \\
QT $k{=}5$ & 3.3\% & 0.0014 & 0.9125 & 85\% & 0\% & +0.62 & +0.19 \\
QT $k{=}10$ & 0.8\% & 0.0004 & 0.9134 & 45\% & 0\% & +0.76 & +0.24 \\
\midrule
\emph{original} & 0.0\% & 0.0000 & 0.9138 & \multicolumn{4}{c}{} \\
\bottomrule
\end{tabular}

%% file: tables/quality.tex
\begin{tabular}{@{}l r rrr r@{}}
\toprule
 & KernelSHAP & \multicolumn{4}{c}{LIME weighted local $R^2$} \\
\cmidrule(lr){2-2}\cmidrule(lr){3-6}
Condition & max $|\varepsilon_{\mathrm{recon}}|$ & median & IQR & \% $<0.5$ & median ($\mathtt{disc}{=}$True) \\
\midrule
original & 1.2e-16 & 0.545 & [0.518, 0.606] & 20.3\% & 0.623 \\
EW $k{=}3$ & 1.1e-16 & 0.544 & [0.517, 0.602] & 20.3\% & 0.620 \\
EW $k{=}5$ & 1.1e-16 & 0.543 & [0.518, 0.604] & 20.3\% & 0.619 \\
EW $k{=}10$ & 2.2e-16 & 0.544 & [0.518, 0.605] & 20.3\% & 0.620 \\
QT $k{=}3$ & 1.1e-16 & 0.544 & [0.518, 0.601] & 20.3\% & 0.624 \\
QT $k{=}5$ & 1.3e-16 & 0.544 & [0.517, 0.603] & 20.3\% & 0.624 \\
QT $k{=}10$ & 1.2e-16 & 0.545 & [0.519, 0.605] & 20.3\% & 0.623 \\
\bottomrule
\end{tabular}

%% file: sections/discussion.tex
\section{Discussion}
\label{sec:discussion}

\subsection{Magnitude and ranking are different kinds of instability}

The clearest pattern in our results is a dissociation. Attribution
\emph{magnitudes} move substantially under discretization---for KernelSHAP, by
up to \ShapRatioMax{} times the explainer's own sampling noise---while the
feature-importance \emph{ranking} is almost untouched, with median Spearman
correlation never below \SpearMin{} and top-3 overlap of \TopThreeMin{}
throughout. These are not two measurements of one phenomenon.

The practical reading depends on how an explanation is consumed. A reader who
uses a local explanation as a shortlist---``which three features mattered
most?''---would observe essentially nothing across any of our conditions. A
reader who uses the magnitudes---to compare the contribution of age against
that of occupation, to decide whether age mattered ``a lot'' or ``a little'',
or to justify a decision to an affected person---would be reading numbers that
shifted materially under a preprocessing choice that left held-out accuracy
unchanged. The second use is the one that motivates local attribution in the
first place.

This also means the common practice of reporting explanation stability as a
single rank-correlation number can mask the effect entirely. Our Spearman
values would support a confident claim that the explanations are stable. The
magnitude results would not.

\subsection{What the noise floor changes}

Measuring each explainer against its own re-run variability changed the
qualitative conclusion for LIME, not merely its precision. LIME's raw
attribution change under discretization is non-zero in every condition and
would be reported as an effect; compared against its own sampling noise, it
exceeds the floor in only \LimeNAboveNoise{} of \LimeNCond{} conditions on the
raw scale. At QT $k{=}5$ it cannot be distinguished from the floor, and in the
other $k \geq 5$ conditions it is \emph{smaller} than what re-running the
explainer produces. The ranking comparison points the same way: for LIME, two
repeats of the \emph{unchanged} condition disagree more than the original and
discretized conditions do. Both comparisons pair the discretized condition with
condition A at a shared seed, so they show that LIME's representation effect is
small relative to its run-to-run variation, not that the effect is absent.

This supports a methodological recommendation that is cheap to follow. Any
claim that some pipeline choice destabilizes an explanation should be
accompanied by the movement that repeating the explainer produces under no
change at all. Both quantities come from the same code path, and the comparison
costs only additional seeds. \citet{zhang2019lime} characterized LIME's
sampling uncertainty qualitatively; our contribution here is simply to use it
as the denominator.

\subsection{Why the two explainers diverge}

Three non-exclusive accounts are consistent with our data, and our design
cannot distinguish them.

First, a \emph{definitional} account. Shapley values satisfy
$\sum_j \phi_j = f(x) - \mathbb{E}[f]$, so any change in the model output must
be redistributed across features. A LIME coefficient is an unconstrained local
slope, and a shift in output level can be absorbed by the surrogate's
intercept. The correlation between $\dphi$ and $\dpabs$ is
\RhoShapMin{}--\RhoShapMax{} for SHAP but about \RhoLimeMed{} for LIME, which
is what this account predicts.

Second, a \emph{structural} account, following from
Section~\ref{sec:what-is-explained}. For KernelSHAP the quantizing wrapper is
exactly equivalent to explaining the unchanged model at a quantized instance
against a quantized background: the discretization is fully internalized into
both the instance and the reference distribution. For LIME no such equivalence
holds---it fits a linear surrogate in the un-quantized coordinates to a
function that is piecewise constant in age. A linear fit over a neighbourhood
spanning several bins will tend to average across the steps, which would
attenuate exactly the signal we are trying to measure.

Third, a \emph{noise} account: LIME's higher sampling variability
(\SdRatioMin{}--\SdRatioMax{}$\times$ SHAP's within-instance spread) may simply
mask a real effect of similar size. These accounts have different
implications---the second and third suggest LIME's apparent robustness is an
artefact of its estimator rather than a property worth relying on---and
separating them is the most valuable follow-up we can identify.

\subsection{Relation to prior findings}

Our SHAP results are consistent with \citet{hwang2025shap}: discretizing a
continuous feature shifts SHAP attributions, including when the classifier is
held fixed. We add that
the shift survives comparison against KernelSHAP's own estimation noise at
practical budgets, which matters given the variance documented by
\citet{covert2021kernelshap}; that it is accompanied by near-perfect prediction
agreement; and that the corresponding effect for LIME is, at most bin counts,
no larger than that explainer's own run-to-run variability.

Our fidelity result instantiates the separation argued by
\citet{yeh2019infidelity}: LIME's median weighted local $R^2$ stays between
\LimeRsqMin{} and \LimeRsqMax{} across all conditions, including the one in
which its age attribution moved beyond its own noise floor, so the median
provides almost no signal about the change. That matters because local $R^2$ is the
quality number LIME reports by default and is often the only one a user sees.

The disagreement we observe is of a different type from that catalogued by
\citet{krishna2024disagreement}: theirs is between methods on one input, ours is
within one method across representations of one input. Both leave a consumer
without a principled way to tell a meaningful difference from an artefact, and
our results suggest the within-method case can be addressed---at least
partially---by reporting a noise floor.

\subsection{Implications for practitioners}

Three implications follow directly from the measurements, and we state them no
more strongly than the evidence supports. (i) Preprocessing choices belong in
the provenance of an explanation: an attribution magnitude is a statement about
a representation, not only about a feature. (ii) Checking that held-out
accuracy survived a pipeline change says little about whether the
per-instance explanations survived it---in our experiment ROC-AUC moved by at
most \AUCDrop{} while most instances' attributions moved beyond the
explainer's noise. (iii) LIME's \discflag{} default silently imposes a
representation of its own; any study of feature representation must disable it
or account for it, and users comparing LIME explanations across pipelines
should check which setting was in force.

We deliberately do not conclude that one explainer is more trustworthy.
KernelSHAP responds more to the representation change and is more consistent
across repeats; LIME responds less and varies more. Whether responsiveness here
is a virtue (faithfully tracking a genuinely changed effective input) or a
defect (instability under an irrelevant pipeline detail) depends on what the
user believes the explanation is \emph{about}, which is a question our
experiment does not answer.

%% file: sections/limitations.tex
\section{Limitations and Threats to Validity}
\label{sec:limitations}

\paragraph{The representation change and the effective-input change cannot be
separated.} This is the central limitation and it is structural, not a
shortcoming of execution. Section~\ref{sec:what-is-explained} establishes that
for KernelSHAP the frozen-model condition \emph{is} the unchanged model
explained at a quantized instance against a quantized background. An
attribution change therefore has two inseparable sources: the explainer
responding to a different representation, and the model genuinely behaving
differently on a different input. We mitigate this by reporting prediction
agreement on every comparison and by making H3 about the coupling itself, but
no frozen-model design can remove it. All attribution results should be read as
conditional on the reported prediction changes. Establishing a ``pure
explainer effect'' would require a manipulation that provably leaves
$f$ invariant on the support of the perturbation distribution, which
discretization does not.

\paragraph{One dataset, one feature, one model family.} We study a single
continuous feature (\texttt{age}) in a single dataset (UCI Adult) with a single
model family (random forest). Nothing here establishes that the pattern
generalizes to other features, datasets or model classes. The design is a
controlled probe, not a survey. The quantitative values---ratios to the noise
floor, the $k$ at which LIME's curve crosses its floor---are specific to this
setting and should not be transferred.

\paragraph{The dataset itself.} UCI Adult is a 1994 census extract with a coarse
race encoding, a binary sex field, an arbitrary income threshold and documented
associations between the label and protected attributes
\citep{ding2021retiring}. We use it as a methods benchmark for comparing
explainers. No conclusion in this paper is a claim about income, about people,
or about fairness.

\paragraph{Both explainers evaluate the model off the data manifold.}
KernelSHAP's coalition rows splice the instance with background rows and LIME's
perturbations sample each feature's marginal independently, so both query the
model at combinations that are rare or impossible---a 17-year-old with a
doctorate, say. We use the libraries' default marginal formulations. As
\citet{aas2021dependent} and \citet{chen2020true} discuss, conditional
formulations answer a different question and would plausibly give different
answers here; part of the SHAP--LIME divergence may reflect how each constructs
its reference rather than sensitivity as such. Supplying
\texttt{categorical\_features} to LIME keeps perturbed rows on the one-hot
simplex, but the category is still drawn from its marginal independently of the
other features; this is a mitigation, not a fix.

\paragraph{Finite explainer budgets.} We use 1500 of $2^{12} = 4096$ possible
coalitions and 40 background rows for KernelSHAP, and 5000 perturbations for
LIME. The seed-noise floor quantifies the resulting variability \emph{at these
budgets}; a different budget would move both the floor and the measured effect,
and \citet{covert2021kernelshap} show KernelSHAP's variance depends strongly on
the sample size. Our conclusions are stated relative to the recorded budgets
and should not be read as budget-free.

\paragraph{The SHAP--LIME magnitude comparison is a comparison of normalized
quantities, not of equivalent ones.} A SHAP value is in probability units and
the values sum to $f(x) - \mathbb{E}[f]$; a LIME coefficient, with
\discflag{} to \texttt{False}, is a ridge slope per training standard
deviation, and for a categorical feature it is per ``same level as the
instance'' indicator. Normalizing to shares of total attribution mass makes the
magnitudes \emph{comparable}, not \emph{equivalent}. As
Section~\ref{sec:discussion} notes, the efficiency constraint may make part of
SHAP's larger response definitional. Rank-based metrics do not suffer from
this, and the ranking results are correspondingly more directly comparable.

\paragraph{Local fidelity is not a validity check.} Neither quality diagnostic
we report measures whether an explanation is correct. KernelSHAP's
reconstruction error is exact by construction once $\ell_1$ regularization is
disabled, so it carries no information. LIME's median weighted local $R^2$
barely changed under a manipulation that, at the coarsest binning, moved its
attributions beyond its own noise floor. Neither should be used to
certify an explanation, and our use of them is purely descriptive.

\paragraph{Stability is not correctness.} The whole paper measures how
explanations \emph{change}, never whether any of them is right. An explainer
that returned a constant vector would be perfectly stable. Our results
constrain which explanations can be relied upon to be reproducible under a
preprocessing change; they say nothing about which, if either, faithfully
describes the model's decision.

\paragraph{Sample size and uncertainty.} \NInstances{} evaluation instances is
a small sample of the \NTest{}-row test partition, chosen for compute reasons. All
intervals are cluster bootstraps that resample whole instances, so they reflect
this, but rare instance types may be absent entirely and the intervals on
per-condition medians are correspondingly wide. The paired design gives
reasonable power for within-instance comparisons and much less for statements
about the test distribution as a whole.

\paragraph{Discretization design choices.} Collapsing a bin to its midpoint is
one of several defensible choices; the mean, the mode, or retaining a bin index
would behave differently. The midpoint-versus-training-median sensitivity run
bounds this at $k=5$ but does not survey the space. Separately, age is
integer-valued, so quantile edges can tie and the requested bin count may not
be achievable; the implementation collapses duplicate edges and records the
effective count. This did not trigger for the reported $k$ values, but it
changes what ``$k=10$'' would mean on a coarser feature.

\paragraph{The secondary experiment is confounded by construction.} Retraining
on discretized data changes the model \emph{and} the representation together.
We report it as a contrast and keep it in separate tables; it is not a
decomposition and cannot be used to attribute the difference to either cause.

\paragraph{No causal claims.} The design supports statements of the form
``under this model, this discretization and these budgets, attributions moved by
this much relative to this noise floor''. It does not support claims about
\emph{why} one explainer responds more than the other; the three accounts in
Section~\ref{sec:discussion} remain unseparated.

%% file: sections/conclusion.tex
\section{Conclusion}
\label{sec:conclusion}

We asked how raw versus discretized representations of a continuous feature
affect local SHAP and LIME explanations under a fixed prediction model, and
whether the two explainers differ. Within the scope of one dataset, one
feature, one model family and the recorded budgets, the answer has four parts.

First, the question needs restating before it can be answered. For KernelSHAP,
a frozen model behind a quantizing wrapper is exactly equivalent to the
unchanged model explained at the quantized instance against a quantized
background (verified to machine precision). The comparison is between two
different effective inputs; it is not a pure explainer effect, and no
frozen-model design makes it one. For LIME the equivalence fails, because its
perturbations are drawn in the raw coordinates---so the two explainers are not
even solving the same transformed problem.

Second, discretization moves KernelSHAP's age attribution beyond that
explainer's own sampling noise in all \ShapNCond{} pre-specified conditions
(median ratios \ShapRatioMin{}--\ShapRatioMax{}), whereas LIME's exceeds its
noise floor in only \LimeNAboveNoise{} of \LimeNCond{} on the raw scale. At
$k \geq 5$ LIME's response is no larger than the variation between two of its
own runs: indistinguishable from it at QT $k{=}5$, and smaller than it in the
other three conditions. SHAP responds
more (\MagRatioMin{}--\MagRatioMax{}$\times$) and varies less
(\SdRatioMin{}--\SdRatioMax{}$\times$ smaller within-instance spread).

Third, attribution instability and prediction agreement are distinct.
Predicted classes changed for at most \FlipMax{}\% of instances and held-out
ROC-AUC never fell below \AUCWorst{}, yet up to \ShapNoFlipMax{}\% of instances
carried a large SHAP attribution change with no class flip.

Fourth---and this is a null result we report as one---feature-importance
\emph{rankings} barely move. Median Spearman correlation stays above
\SpearMin{} and top-3 overlap is \TopThreeMin{} in every condition. For LIME,
re-running the explainer perturbs the ranking more than discretizing the
feature does (a comparison made at a shared seed against a floor measured
across seeds). The instability we measure lives in attribution magnitudes, not
in which feature matters most.

We draw no conclusion about which explainer is preferable. We do draw one
methodological conclusion: a representation-sensitivity result reported without
the explainer's own re-run variability may report an effect that is no larger
than the difference a user would see from simply re-running the explainer, and
in this experiment that was the case for LIME at $k \geq 5$.

\paragraph{Future work.} Four experiments would materially strengthen or
qualify these findings, in descending order of value. (i)~\emph{Separating the
three accounts of the SHAP--LIME divergence} (Section~\ref{sec:discussion}) by
sweeping LIME's kernel width: the structural account predicts the effect
emerges as the neighbourhood narrows below the bin width, the noise account
does not. (ii)~\emph{Repeating the design with a dependence-aware reference
distribution} \citep{aas2021dependent,chen2020true}, which would test whether
the divergence reflects how each method constructs its reference rather than
sensitivity as such. (iii)~\emph{Extending to several datasets, several
continuous features and at least one non-tree model}, which is the only way to
establish that any of this generalizes. (iv)~\emph{A human-subjects study} of
whether the mid-ranking magnitude instability we measure is large enough to
change a reader's decision---our results quantify the change but say nothing
about whether it matters to the people explanations are produced for.